\documentclass[sigconf]{acmart}

\usepackage{booktabs} 

\usepackage{algorithm}  
\usepackage{algorithmic}  
\usepackage{subfigure}
\usepackage{rotating}
\usepackage{multirow}
\usepackage{tablefootnote}

\newcommand{\ep}{\mathbb{E}}

\setcopyright{acmcopyright}

\begin{document}
\copyrightyear{2018} 
\acmYear{2018} 
\setcopyright{acmcopyright}
\acmConference[KDD '18]{The 24th ACM SIGKDD International Conference on Knowledge Discovery \& Data Mining}{August 19--23, 2018}{London, United Kingdom}
\acmBooktitle{KDD '18: The 24th ACM SIGKDD International Conference on Knowledge Discovery \& Data Mining, August 19--23, 2018, London, United Kingdom}
\acmPrice{15.00}
\acmDOI{10.1145/3219819.3220047}
\acmISBN{978-1-4503-5552-0/18/08}

\title{Learning-to-Ask: Knowledge Acquisition via {20 Questions}}

\author{Yihong Chen}
\authornote{This work was mainly done when the authors were at Microsoft Research.}
\affiliation{
  \institution{Dep. of Electr. Engin. \\Tsinghua University, Beijing, China}
}
\email{cyh16@mails.tsinghua.edu.cn}

\author{Bei Chen}
\affiliation{
  \institution{Microsoft Research}
  \city{Beijing, China} 
}
\email{beichen@microsoft.com}

\author{Xuguang Duan}
\authornotemark[1]
\affiliation{
  \institution{Dep. of Electr. Engin. \\Tsinghua University, Beijing, China}
}
\email{dxg14@mails.tsinghua.edu.cn}

\author{Jian-Guang Lou}
\affiliation{
  \institution{Microsoft Research}
  \city{Beijing, China} 
}
\email{jlou@microsoft.com}

\author{Yue Wang}
\affiliation{
  \institution{Dep. of Electr. Engin. \\Tsinghua University, Beijing, China}
}
\email{wangyue@mail.tsinghua.edu.cn}

\author{Wenwu Zhu}
\affiliation{
  \institution{Dep. of Comp. Sci. \& Technol. Tsinghua University, Beijing, China}
}
\email{wwzhu@mail.tsinghua.edu.cn}

\author{Yong Cao}
\authornotemark[1]
\affiliation{
  \institution{Alibaba AI Labs}
  \city{Beijing, China} 
}
\email{yohncao.cy@alibaba-inc.com}

\renewcommand{\shortauthors}{Y. Chen, B. Chen, X. Duan, J. Lou, Y. Wang, W. Zhu, and Y. Cao}

\begin{abstract}
Almost all the knowledge empowered applications rely upon accurate knowledge, which has to be either collected manually with high cost, or extracted automatically with unignorable errors. In this paper, we study \textit{20 Questions}, an online interactive game where each question-response pair corresponds to a fact of the target entity,  to acquire highly accurate knowledge effectively with nearly zero labor cost. Knowledge acquisition via \textit{20 Questions} predominantly presents two challenges to the intelligent agent playing games with human players. The first one is to seek enough information and identify the target entity with as few questions as possible, while the second one is to leverage the remaining questioning opportunities to acquire valuable knowledge effectively, both of which count on good questioning strategies. To address these challenges, we propose the \textit{Learning-to-Ask} (LA) framework, within which the agent learns smart questioning strategies for information seeking and knowledge acquisition by means of deep reinforcement learning and generalized matrix factorization respectively. In addition, a Bayesian approach to represent knowledge is adopted to ensure robustness to noisy user responses. Simulating experiments on real data show that LA is able to equip the agent with effective questioning strategies, which result in high winning rates and rapid knowledge acquisition. Moreover, the questioning strategies for information seeking and knowledge acquisition boost the performance of each other, allowing the agent to start with a relatively small knowledge set and quickly improve its knowledge base in the absence of constant human supervision.
\end{abstract}

%
%



\maketitle

\section{Introduction}
There is no denying that knowledge plays an important role in many intelligent applications including, but not limited to, question answering systems, task-oriented bots and recommender systems. However, it remains a persistent challenge to build up accurate knowledge bases effectively, or rather, acquire knowledge at affordable cost. Automatic knowledge extraction from existing resources like semi-structured or unstructured texts~\cite{tandon2014webchild} usually suffers from low precisions in spite of high recalls. Manual knowledge acquisition, while providing very accurate results, is notoriously costly. Early work explored crowdsourcing to alleviate the manual cost. To motivate user engagement, the knowledge acquisition process was transformed into enjoyable interactive games between the game agents and users, termed as \textit{games with a purpose} (GWAP)~\cite{von2008designing}. Drawing inspirations from GWAP, we find that the spoken parlor game, \textit{20 Questions}\footnote{https://en.wikipedia.org/wiki/Twenty\_Questions}, is an excellent choice to be equipped with the purpose of knowledge acquisition, via which accurate knowledge can be acquired with nearly zero labor cost.

\begin{figure}
\includegraphics[height=2.2in, width=3.35in]{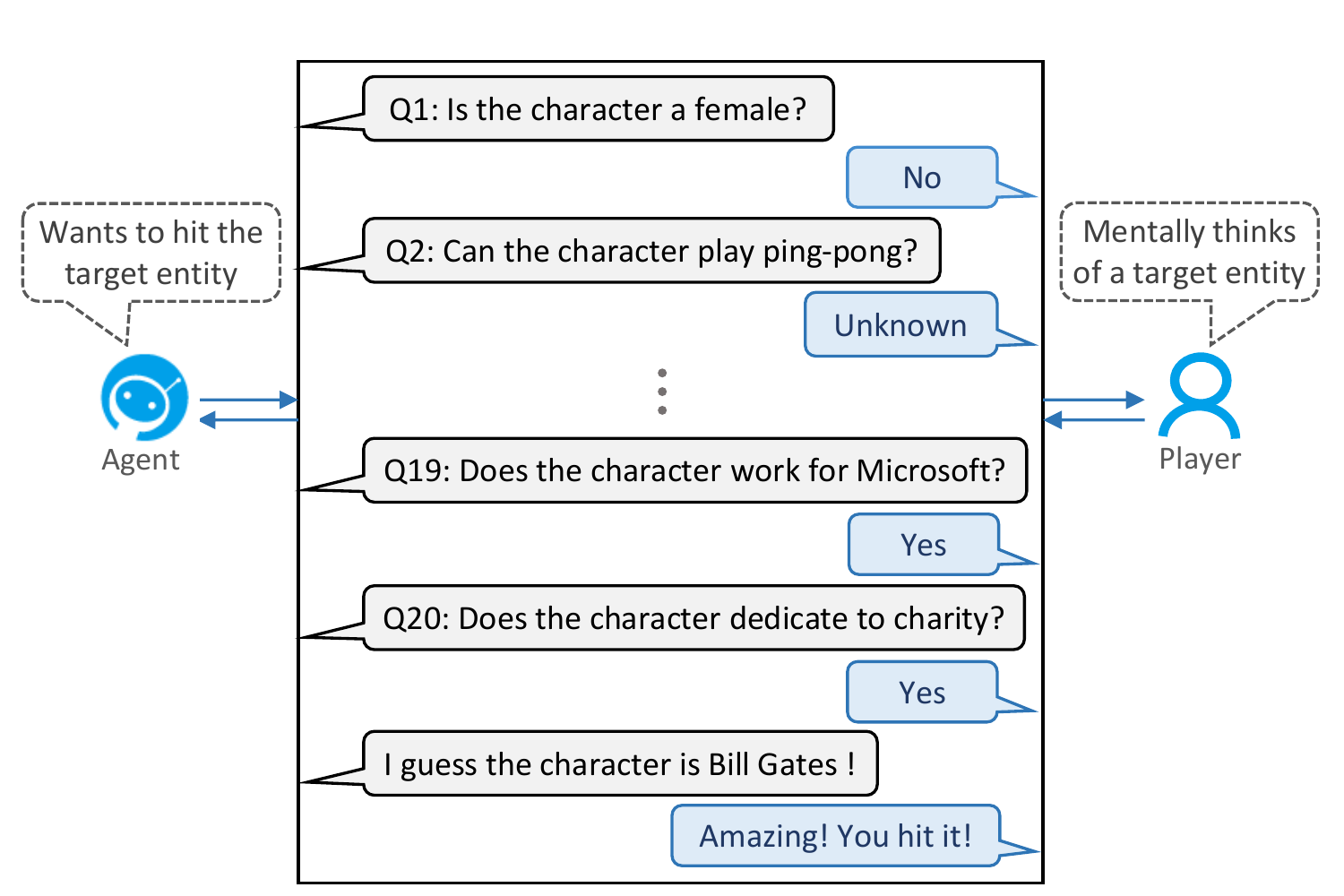}
\caption{An example episode of \textit{20 Questions}.}
\label{fig:game flow}
\end{figure}

In the online interactive \textit{20 Questions}, the game agent plays the role of the guesser and tries to figure out what is in the mind of the human player. Figure \ref{fig:game flow} gives an example episode. 
At the beginning, the player thinks of a target entity mentally. Then, the agent asks yes-no questions, collects responses from the player, and finally tries to guess the target entity based on the collected responses. The agent wins the episode if it hits the target entity. Otherwise, it fails. 
In essence, \textit{20 Questions} presents to the agent an information-seeking task~\cite{marchionini1997information,wilson2000human}, which aims at interactively seeking information about the target entity. During a \textit{20 Questions} episode, each question-response pair is equivalent to a fact of the target entity. For example, <"Does the character work for Microsoft?", "Yes"> with target "Bill Gates" corresponds to the fact "Bill Gates works for Microsoft.", and <"Is the food spicy?", "No"> with target "Apple" corresponds to the fact "Apple is not spicy.". 
Hence, in addition to seeking information using existing knowledge for game playing, the agent can actually collect new knowledge by asking questions without existing answers on purpose. Since the player does not know whether the current question is for information seeking or for knowledge acquisition, he/she feels nothing different about the game while contributing his/her knowledge unknowingly. What's more, the wide audience of \textit{20 Questions} makes it an excellent carrier to incorporate human contributions into the knowledge acquisition process. After a reasonable number of game sessions, the agent's knowledge becomes stable and accurate as the acquired answers are voted across many different human players. 

However, integrating knowledge acquisition into the original information-seeking \textit{20 Questions} poses two serious challenges for the agent. 
(1) The agent needs efficient and robust \textit{information seeking} (IS) strategies to work with noisy user responses and hit the target entity with as few questions as possible.
As the questioning opportunities are limited, the agent earns more knowledge acquisition opportunities if it can identify the target entity with less questions. Hitting the target entity accurately is not only the precondition for knowledge acquisition, since the facts have to be linked to the correct entity, but also the key to attract human players because nobody wants to play with a stupid agent.
(2)  To acquire valuable facts, the agent needs effective \textit{knowledge acquisition} (KA) strategies which can identify important questions for a given entity. Due to the great diversity in entities, most questions are not tailored or even irrelevant to a given entity and they shouldn't be asked during the corresponding KA. For example, KA on "Bill Gates" should start with meaningful questions rather than "Did the character play a power forward position?", which is more suitable for KA on basketball players.

In this paper, to address the challenges, we study effective IS and KA strategies under a novel \textit{Learning-to-Ask} (LA) framework. Our goal is to learn both IS and KA strategies which boost each other, allowing the agent to start with a relatively small knowledge base and quickly improves in the absence of constant human supervision. Our contributions are summarized as follows:
\begin{itemize}
\item We are the first to formally integrate IS strategies and KA strategies into a unified framework, named Learning-to-Ask, which turns the vanilla \textit{20 Questions} game into an effective knowledge harvester with nearly zero labor cost.
\item Novel methods combining reinforcement learning (RL) with Bayesian treatment are proposed to learn IS strategies in LA, which allow the agent to overcome noisy responses and hit the target entity with limited questioning opportunities.
\item KA strategies learning in LA is achieved with a generalized matrix factorization (GMF) based method so that the agent can infer the most important and relevant questions given the current target entity and knowledge base status.
\item We conduct experiments against simulators built on real world data. The results demonstrate the effectiveness of our methods with high-accuracy IS and rapid KA.
\end{itemize}

\textbf{Related Work.} 
Knowledge acquisition has been widely studied. Early projects like Cyc~\cite{lenat1995cyc} and OMCS~\cite{singh2002open} leveraged crowdsourcing to build up the knowledge base. GWAP were proposed to make knowledge acquisition more attractive to users. Games were designed for a dual purpose of entertainment and completing knowledge acquisition tasks: ESP game \cite{von2004labeling} and Peekaboom \cite{von2006peekaboom} for labeling images; Verbosity \cite{von2006verbosity}, Common Consensus \cite{lieberman2007common}, and a similar game built on a smartphone spoken dialogue system \cite{Otani2016Large-ScaleSystem} for acquiring commonsense knowledge. 
As for \textit{20 Questions}, it is originally a spoken parlor game that has been extensively explored in a variety of disciplines. \textit{20 Questions} was used as a research tool in cognitive science and developmental psychology \cite{Courage1989ChildrensQuestions}, also powered a commercial product\footnote{http://www.20q.net}. Previous work \cite{zhao2016towards} has used reinforcement learning for \textit{20 Questions}, where the authors treated the game as a task-oriented dialog system and applied hard-threshold guessing with the assumption that the responses from players were completely correct except for language understanding errors. In contrast, we explicitly consider the uncertainty in the players' feedback and equip the game with the purpose of knowledge acquisition.
Although \textit{20 Questions} was proposed for commonsense knowledge acquisition in ~\cite{Speer2008AnModel}, the work predominantly focused on designing an interface specially for collecting facts in OMCS, using statistical classification methods combined with fine-tuned heuristics. We instead study the problem of learning questioning strategies for both IS and KA, which is more meaningful in terms of powering a competitive agent to attract the players. Moreover, as our methods do not rely on lots of handcrafted rules or specific assumptions about the knowledge bases, they can be generalized to other knowledge bases with little pain, generating questioning strategies which gradually improve as game sessions increase. To the best of our knowledge, we are the first to incorporate RL and GMF into a unified framework to learn questioning strategies, which can be used by the agents for the dual purpose of playing entertaining \textit{20 Questions}, i.e. a game of information-seeking nature, and acquiring new knowledge from human players. 

The rest of the paper is organized as follows. Section 2 introduces LA, the proposed unified framework for questioning strategies generation. Multiple methods to implement LA are presented in Section 3 and empirically evaluated in Section 4, followed by conclusion and discussion in Section 5.

\section{Learning-to-Ask Framework}
In this section, we present \textit{Learning-to-Ask} (LA), a unified framework under which the agent can learn both information seeking and knowledge acquiring strategies in \textit{20 Questions}. To make it clear, we first introduce the knowledge base which backs up our agent and then we present LA in detail. 

\begin{figure}
\includegraphics[height=1.35in, width=3.2in]{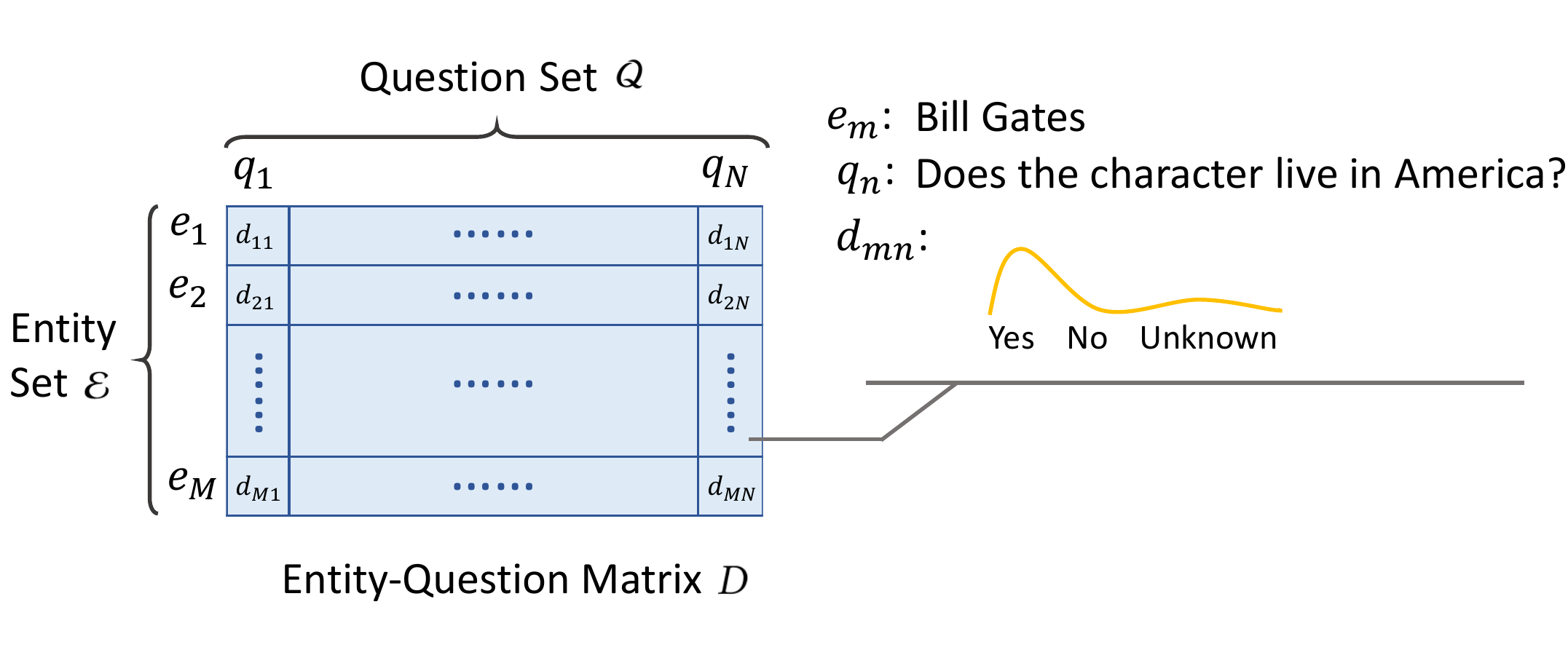}
\caption{The knowledge base: entity-question matrix.}
\label{fig:kb2matrix}
\end{figure}

As shown in Figure \ref{fig:kb2matrix}, the knowledge base is represented as a $M \times N$ entity-question matrix $D$, with each row and each column corresponding to an entity in $\mathcal{E}=\{e_1, e_2, ..., e_M\}$ and a question in $\mathcal{Q}=\{q_1, q_2, ..., q_N\}$ respectively. An entry $d_{mn}$ in $D$ stores the agent's knowledge about the pair $(e_m, q_n)$. Although it may be tempting to regard 'Yes' or 'No' as knowledge for the pair $(e_m, q_n)$, the deterministic approach suffers from human players' noisy responses (They may make mistakes unintentionally or give 'Unknown' responses). We thus resort to a Bayesian approach, modelling $d_{mn}$ as a random variable which follows a multinoulli distribution over a possible set of responses $\mathcal{R}=$ \{ \textit{yes, no, unknown} \}. The exact parameters of $d_{mn}$ can be estimated directly from game logs or replaced by a uniform distribution if the entry is missing from the logs. Notably, for an initial automatically constructed knowledge base, $d_{mn}$ can also be roughly approximated even though there is no game logs yet. Hence, our methods actually overcome the cold-start problem encountered in most knowledge base completion scenarios.

\begin{figure}
\includegraphics[height=1.85in, width=3in]{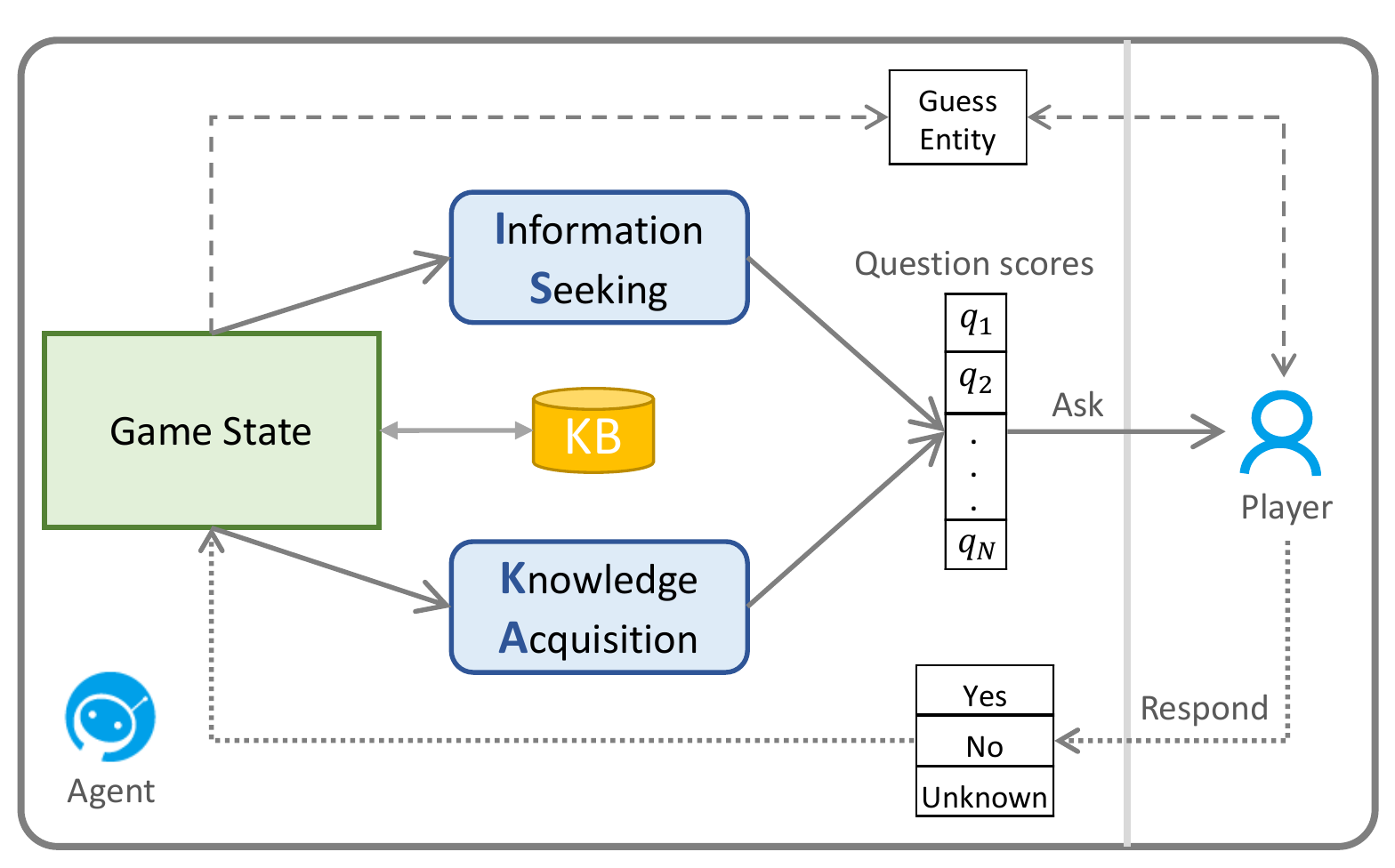}
\caption{Learning-to-Ask framework.}
\label{fig:framework}
\end{figure}

Backed by the knowledge base described above, the agent learns to ask questions within the LA framework. Figure \ref{fig:framework} summarizes the LA framework. At each step in an episode, the agent manages  the game state, which contains the history of questions and responses over the past game steps besides the guess. Two modules, termed as IS Module and KA Module, are devised to generate the questions for the dual purposes of IS and KA. Both modules output scores for the candidate questions, from which the agent chooses the most appropriate one to ask the human player. The response given by the human player is used to update the game state. At the end of each episode, the agent sends its guess to the human player for the final judgment. Through experiences of such episodes, the agent gradually learns the questioning strategies that identify the target entity effectively and improves its own knowledge base. The detailed functionalities of the modules are listed below.

\textbf{IS Module.} This module generates questioning strategies to collect necessary information to hit the target entity at the final guess. At each game step $t$, IS module detects the ambiguity in the collected information $(a_1, x_1, a_2, x_2, ..., a_{t-1}, x_{t-1})$, where $a$ denotes a question and $x$ denotes a response. And it scores the candidate questions according to how much uncertainty they can resolve about the target entity. The agent asks the next question based on the scores. Once adequate information collected, the agent makes a guess $g$. 

\textbf{KA Module.} This module generates questioning strategies to improve the agent's knowledge base i.e. acquire knowledge about the missing entries in $D$. We assume that the ideal knowledge base for the agent should be consistent with the one human players possesses. Hence, the most valuable missing entries that KA module aims to acquire are those known to the human players but unknown to the agent. KA module scores the candidate question based on the values of entries pertaining to the guess $g$. The agent asks the question according to the scores and accumulates the response into its knowledge buffer.

Importantly, the questioning opportunities are limited. The more opportunities allocated to KA module, the more knowledge the agent acquires. However, the IS module needs enough questioning opportunities as well. Otherwise, the agent may fail to collect adequate information and make a correct guess. Therefore, it is a big challenge to improve both modules in LA. Note that the whole LA framework is invisible to the human players. They simply regard it as an enjoyable game and contribute their knowledge unknowingly while the agent gradually improves itself.

\section{Methods}
So far we have introduced LA, a unified framework under which the agent learns questioning strategies to interact with players effectively. Here we present our methods to implement LA. As described above, the key to implement LA is to instantiate the IS module and the KA module. We leverage deep reinforcement learning to instantiate the IS module, while generalized matrix factorization is incorporated into the KA module. We start by explaining the basic methodology underlying our methods, then we describe the implementation details.   

\subsection{Learning-to-Ask for Information Seeking}
IS module aims to collect necessary information to identify the target entity while the agent interacts with the human player, which is essentially a task of interactive strategy learning.  In particular, no direct instructions are given to the agent and the consequence of asking some question plays out only at the end of the episode. This setting of interactive strategy learning naturally lends itself to reinforcement learning (RL) \cite{Sutton1998ReinforcementIntroduction}. One of the most popular application domains of reinforcement learning is task-oriented dialog systems\cite{li2016deep,gavsic2013line}. Zhao et al.\cite{zhao2016towards} proposed a variant of deep recurrent Q-network (DRQN) for end-to-end dialog state tracking and management, from which we drew inspiration to design our agent. Unlike approaches with subtle control rules and complicated heuristics, the RL-based methods allow the agent to learn questioning strategies solely through experience. At the beginning, the agent has no idea what effective questioning strategies should be. It gradually learns to ask "good" questions by trial and error. Effective questioning strategies are reinforced by the reward signal that indicates whether the agent wins the game, namely whether it collects enough information to hit the target entity.

Formally, let $T$ be the number of total steps in an episode ($T=20$ in \textit{20 Questions}),  and $T_1 < T$ be the number of questioning opportunities allocated to the IS module. At step $t$, the agent observes the current game state $s_t$, which ideally summarizes the game history over the past steps, and takes an action $a_t$ based on a policy $\pi$. In our case, an action corresponds to a question in the question set $\mathcal{Q}$ and a policy corresponds to a questioning strategy that maps game states to questions. After asking a question, the agent receives a response $x_t$ from the human player and a reward signal $r_t$ reflecting its performance. The primary goal of the agent is to find an optimal policy that maximizes the expected cumulative discounted reward over an episode:
\begin{equation}
\label{eq_rt}
\ep[R_t] = \ep\left[ \sum_{i=t}^{T_1}\gamma^{i-t}r_i \right],
\end{equation}
where $\gamma \in [0,1]$ is a discount factor that determines the importance of future rewards. This can be achieved by Q-learning, which learns a state-action value function or \textit{Q-function} $Q(s, a)$ and constructs the optimal policy by greedily selecting the action with the highest value in each state.

Given that the states in IS are complex and high-dimensional, we resort to deep neural networks to approximate the Q-function. This technique,  termed as $DQN$, was introduced by Mnih et al.\cite{mnih2015human}. DQN utilizes \textit{experience replay} and \textit{target network} to mimic the supervise learning case and improve the stability.  The Q-network is updated by performing gradient descent on the loss \begin{equation}
l(\theta) = \ep_{(s_t, a_t, s_{t+1}, r_t) \sim U}[(r_t + \gamma \max_{a_{t+1}}Q(s_{t+1}, a_{t+1}; \tilde\theta) - Q(s_t, a_t; \theta))^2],
\end{equation}
where $U$ is the experience replay memory that stores past transitions, $\theta$ are the parameters of the behave Q-network and $\tilde\theta$ are the parameters of the target network. Note that $\tilde\theta$ are copied from $\theta$ only after every $C$ updates.  

The above DQN method built upon \textit{Markov Decision Process} (MDP) assumes the states to be fully observable. Although it is possible to manually design such states for IS module by incorporating the whole history, it may sacrifice the learning efficiency of Q-network. To address this problem, we combine embedding technique and deep recurrent Q-network (DRQN) \cite{hausknecht2015deep} which utilizes recurrent neural networks to maintain belief states over the sequence of previous observations. 
Here we first present LA-DQN, a method to instantiate IS module using DQN, and explain its incapability in dealing with large question space and response space. Then we describe LA-DRQN, which leverages DRQN with question and response embedded.

\begin{figure}
\includegraphics[height=1.6in, width=2.6in]{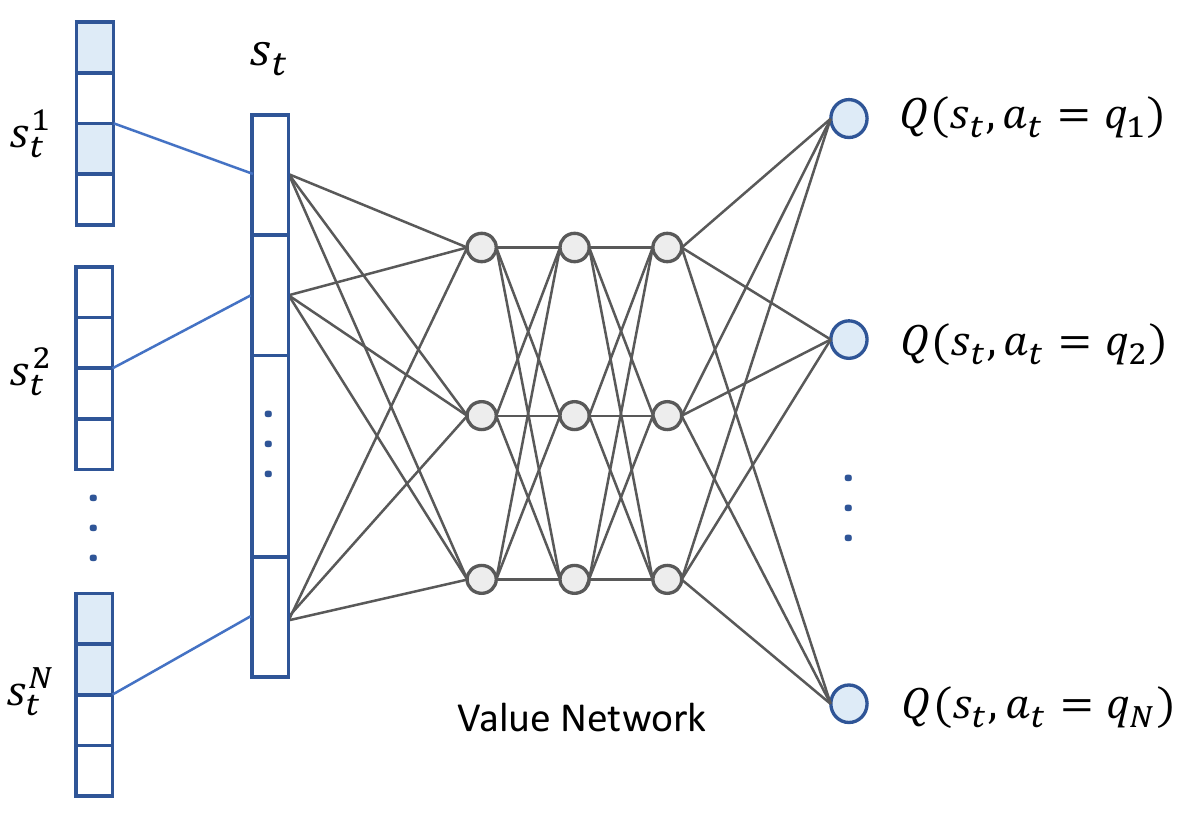}
\caption{Q-network architecture for LA-DQN.}
\label{fig:dqn architecture}
\end{figure}

\subsubsection{LA-DQN}
Let $h_t = (a_1, x_1, a_2, x_2, ..., a_{t-1}, x_{t-1})$ be the history at step $t$, which provides all the information the agent needs to determine the next question. Owing to the required Markov property underlying DQN, a straightforward design for the state $s_t$ is to encapsulate the history $h_t$. Since there are $3$ possible values for responses in our setting (\textit{yes, no, unknown}), the state $s_t$ can be represented as a $4N$-dimensional vector:
\begin{equation}
s_t = [s_{t}^{1}, s_{t}^{2}, ..., s_{t}^{N}]^\top,
\end{equation}
where $s_t^n$ is a $4$-dimensional binary vector associated with the status of the question $q_n \in \mathcal{Q}$ up to step $t$. The first dimension in $s_t^n$ indicates whether $q_n$ has been asked previously i.e whether it appears in the history of the current episode. If $q_n$ has been asked, the remaining $3$ dimensions are one-hot representing the response from player; otherwise, the remaining dimensions are all zeros. The state $s_t$ is feed to the Q-network implemented as Multilayer Perceptrons (MLP) to estimate the state-action values for all questions: $[Q(s_t, a_t = q_1), Q(s_t, a_t = q_2), ..., Q(s_t, a_t = q_N)].$ Figure \ref{fig:dqn architecture} shows the Q-network architecture for LA-DQN. Although there are alternatives for the Q-network architecture, this one allows the agent to obtain the state-action values for all questions in just one forward pass, which greatly reduces computation complexity and improves the scalability of the whole LA framework.

\begin{figure}
\includegraphics[height=1.6in, width=2.91in]{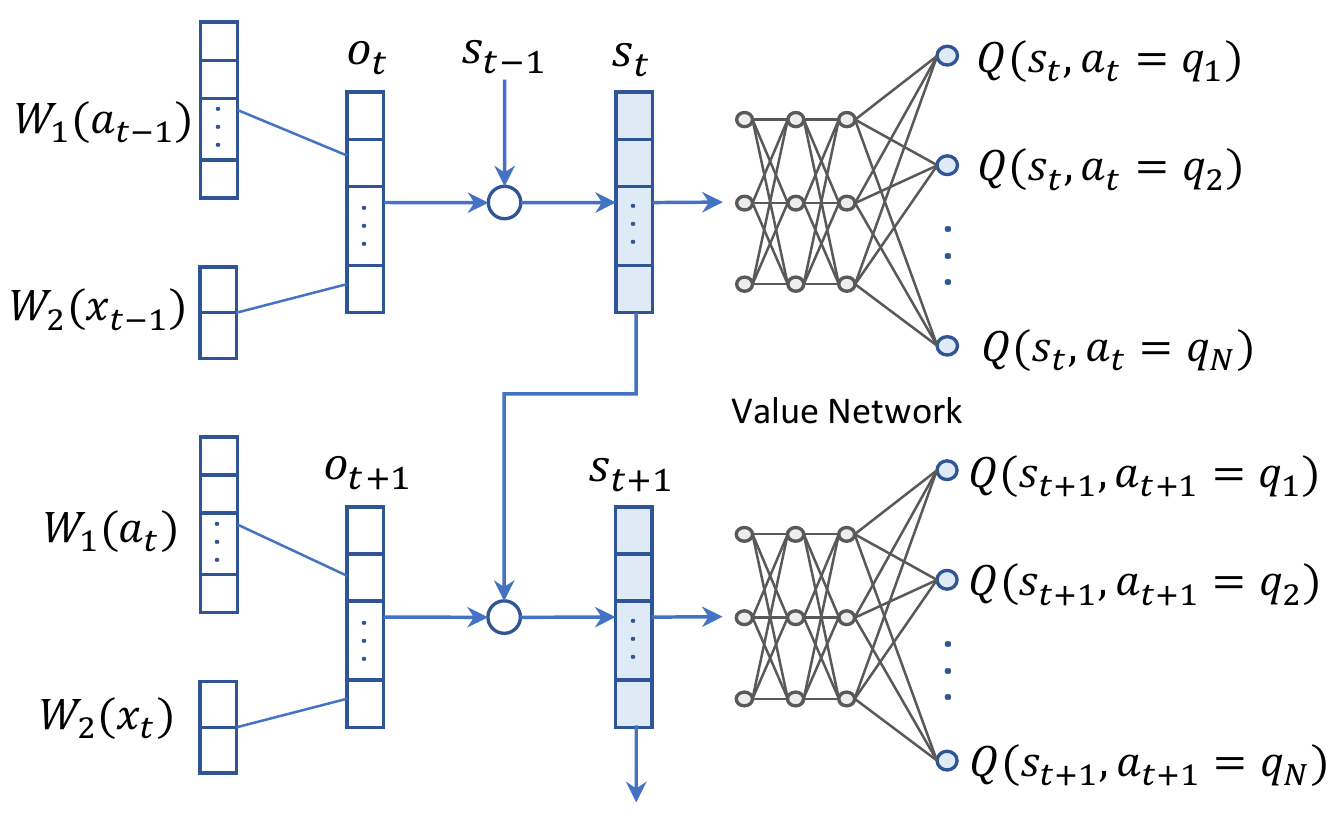}
\caption{Q-network architecture for LA-DRQN.}
\label{fig:drqn architecture}
\end{figure}

\subsubsection{LA-DRQN}
In the LA-DQN method, states are designed to be fully observable, containing all the information needed for the next questioning decision. However, the state space is highly underutilized, and it expands fast as the valid questions and responses increase, which results in slow convergence of Q-Network learning and poor performance of target entity identification. Compared to LA-DQN, LA-DRQN is more powerful in terms of generalization to large question and response spaces.  As shown in Figure \ref{fig:drqn architecture}, dense representations of questions and responses are learned in LA-DRQN. Similar as word embeddings, the question embedding $W_1: q \to \mathbb{R}^{N_1}$ and the response embedding $W_2: x \to \mathbb{R}^{N_2}$ are parametrized functions mapping questions and responses to $N_1$-dimensional real-value vectors and $N_2$-dimensional real-value vectors respectively. The observation at step $t$ is the concatenated embeddings of the question and response at the previous step: 
\begin{equation}
o_t = [W_1(a_{t-1}), W_2(x_{t-1})]^\top.
\end{equation}
To capture information over steps, a recurrent neural network (e.g. LSTM) is employed to summarize the past steps and generate a state representation for current step:
\begin{equation}
s_t = {\rm{LSTM}}(s_{t-1}, o_t),
\end{equation}
where the state $s_t \in \mathbb{R}^{N_3}$. Similar as LA-DQN, the generated state is used as the input to a network to estimate the state-action values for all questions. It is important to realize that complex states bring obstacles to Q-network training while simple states may fail to capture some valuable information. The state design in LA-DRQN combines simple states of concatenated embeddings and recurrent neural networks to summarize the history, which is more appropriate for the IS module. The embedding approach also enables effective learning of model parameters and makes it possible to scale up.

\subsubsection{Guessing via Naive Bayes} 
At step $T_1$, the agent accomplishes the task of IS module and makes a guess on the target entity using the collected responses to the $T_1$ questions. Let $X=(x_1, x_2, ..., x_{T_1})$ denote the response sequence and $g$ denote the agent's guess which can be any entity in $\mathcal{E}=\{e_1,e_2,...,e_M\}$. The posterior of the guess $g$ is
\begin{equation} \label{eq:guess}
P(g|X) \propto  P_0(g) P(X|g) \propto P_0(g) \times \prod_{t=1}^{T_1}P(x_t|g),
\end{equation}
assuming independence among the responses given a specific entity. An ideal choice for the prior $P_0(g)$ is the entity popularity distribution. After computing the posterior probability for each $e_m \in \mathcal{E}$, the agent chooses the one with the maximum posterior probability as the final guess. This Bayesian approach reduces LA's sensibility to the noisy responses given by the players and allows the agent to make a more robust guess.

\subsection{\!Learning-to-Ask for Knowledge Acquisition}
LA framework can make \textit{20 Questions} an enjoyable game with the well-designed IS module described above. Moreover, the incorporation of a sophisticated KA module will make it matter more than just a game.  Successful KA not only boosts the winning rates of the game, but also improves the agent's knowledge base. Intuitively, the agent achieves the best performance when its knowledge is consistent with the human players' knowledge. However, this is usually not the case in practice, where the agent always has less knowledge due to limited resources for knowledge base construction and the dynamic nature of knowledge. Hence, motivating human contributions plays an important part in KA. As described in Section 2, the goal of KA module is to fill in the valuable missing entries. Specifically, given an entity $e_m$, the agent chooses questions from $\mathcal{Q}$ to collect knowledge about the entries $\{d_{mn}\}_{n=1}^N$. Instead of asking questions blindly, we propose LA-GMF, a \textit{generalized matrix factorization} (GMF) based method that takes into account the following two aspects to make KA more efficient.

{\textbf{Uncertainty}}. Consider the knowledge base shown in Figure \ref{fig:kb2matrix}. To fill in an entry $d_{mn}$ means to estimate a multinoulli distribution over possible response values \{\textit{ yes, no, unknown }\} from collected responses. Let $N_{mn}$ denote the number of collected responses for $d_{mn}$. Small $N_{mn}$ leads to high uncertainty of the estimated $d_{mn}$. For this reason, the entries with higher uncertainty should be chosen, so that the agent can refine the related estimations faster. With this in mind, we select $N_c$ questions from $\mathcal{Q}$ to maintain an adaptive candidate question set $\mathcal{Q}_c$, where a question $q_n$ is chosen proportional to $\frac{1}{\sqrt{N_{mn}}}$. Note that, for the missing entries, $N_{mn}$ are initialized as 3, the number of possible response values, since we give them uniform priors. In the cases where the agent receives too many "\textit{unknown}" responses for an entry, the entry should be rejected to be asked any further, because it is of little value and nobody bothers to know about it. 

{\textbf{Value}}. Among the entries with high uncertainty, only some of them are worth acquiring. A valuable question is the one that the human players are more likely to know the answer (They respond with \textit{yes}/\textit{no} rather than \textit{unknown}). Interestingly, given the incomplete $D$, predicting the missing entries is similar to the collaborative filtering in recommendation systems, which utilizes matrix factorization to infer users' preference based on their explicit or implicit adopting history. Recent studies show that matrix factorization (MF) can be interpreted as a special case of neural collaborative filtering~\cite{He2017NeuralFiltering}. Thus, we employ generalized matrix factorization (GMF), which extends MF under the neural collaborative framework, to score entries and rank the candidates in $\mathcal{Q}_c$. To be clear, let $Y \in R^{M \times N}$ be the indicator matrix constructed from $D$: if the entry $d_{mn}$ is known ($N_{mn}\neq 3 $), then $y_{mn} = 1$; otherwise, $y_{mn} = 0$. $Y$ can be factorized as $Y=U\cdot V$, where $U$ is a $M\times K$ real-valued matrix and $V$ is a $K\times N$ real-valued matrix. Let $U_m$ be the $m$-th row of $U$ denoting the latent factor for entity $e_m$, and $V_n$ be the $n$-th column of $V$ denoting the latent factor for question $q_n$. GMF scores each entry $d_{mn}$ and estimates the value as:
\begin{equation}\label{eq:score}
\hat{y}_{mn} = a_{out}(h^\top(U_m\odot V_n)),
\end{equation}
where $\odot$ denotes the element-wise product of vectors, $h$ denotes the weights of the output layer, and $a_{out}$ denotes the non-linear activation function. $U$ and $V$ are learned by minimizing the mean squared error $(y_{mn} - \hat{y}_{mn})^2$.

\begin{algorithm}[tb]
\caption{ One cycle of LA-GMF for KA module.}   
\label{alg:ka}  
\begin{algorithmic}
\STATE {\bfseries Input:} knowledge base $D$, buffer size $N_k$
\STATE construct $Y$ from $D$; factorize $Y=U\cdot V$; $Count = 0$
\WHILE {$Count < N_k$}
\STATE $e_m=g$ is the guess of the current episode
\STATE $\mathcal{Q}_h=\{a_1, a_2, ..., a_{T_1}\}$ is the set of history asked questions
\STATE $\theta_m=(\frac{1}{\sqrt{N_{m1}}}, \frac{1}{\sqrt{N_{m2}}}, ..., \frac{1}{\sqrt{N_{mN}}})$ is the uncertainty parameter
\FOR {$step=1, 2, \dots, T_2$}
\STATE $\mathcal{Q}_c=\varnothing$
\FOR {$i=1, 2, \dots, N_c$}
\STATE $q_i\sim {\rm{Multinomial}}(\theta_m)$ and $q_i \notin \mathcal{Q}_c$ and $q_i \notin \mathcal{Q}_h$
\STATE $\mathcal{Q}_c = \mathcal{Q}_c + q_i$
\ENDFOR
\STATE rank questions in $\mathcal{Q}_c$ by scores using Equation (\ref{eq:score})
\STATE ask the top question $q$ in $\mathcal{Q}_c$, collect the response in buffer
\STATE $\mathcal{Q}_h = \mathcal{Q}_h + q$; $Count = Count + 1$
\ENDFOR
\ENDWHILE
\STATE update $D$ using all the knowledge in buffer
\end{algorithmic}
\end{algorithm}

The collected responses are accumulated into a buffer and periodically updated to the knowledge base. An updating cycle of LA-GMF is shown in Algorithm \ref{alg:ka}, which generally contains multiple episodes. The agent starts with the knowledge base $D$. During an episode, the agent makes a guess $g$ after IS module accomplishes its task in $T_1$ questions, and the remaining $T_2=T-T_1$ questioning opportunities are left for KA module. LA-GMF first samples a set of candidate questions $\mathcal{Q}_c$ and ranks the questions according to Equation \ref{eq:score}. The question with the highest score is chosen to ask the human player. Once the buffer is full, a cycle finishes, and the knowledge base $D$ is updated. Ideally, the improved knowledge base $D$ converges to the knowledge base that the human players possess as the KA process goes on and on. Notably, if the guess provided by the IS module is wrong, the agent fails to acquire correct knowledge in the episode, which we omit in Algorithm \ref{alg:ka}. Therefore, a well-designed IS module is quite important to the KA module, and in return, the knowledge collected by the KA module increases the agent's competitiveness in playing \textit{20 Questions}, making the game more attractive to the human players. Furthermore, in contrast to most works adopting relation inference to complete knowledge base, LA provides highly accurate knowledge base completion due to its interactive nature.

\subsection{Implementation}
In this subsection, we describe details about the implementations of LA-DQN, LA-DRQN and LA-GMF. To avoid duplicated questions which reduce human players' pleasure, we constrain the available questions to those which haven't been asked over the past steps in the current episode. Rewards $r$ for LA-DQN and LA-DRQN are $+1$ for winning, $-1$ for losing, and $0$ for all nonterminal steps. We adopt $\epsilon$-greedy exploration strategy. Some modern reinforcement learning techniques are used to improve the convergence of LA-DQN and LA-DRQN: Double Q-learning \cite{van2016deep} alleviates the overestimation of Q-value and prioritized experience replay \cite{schaul2015prioritized} allows faster backwards propagation of reward information by favoring transitions with large gradients. Since all the possible rewards over an episode fall into $[-1, 1]$, $Tanh$ is chosen as the activation function for the Q-network output layer. Batch Normalization \cite{ioffe2015batch} and Dropout \cite{srivastava2014dropout} techniques are used to accelerate training of Q-networks as well.  As for LA-GMF, we use a $Sigmoid$ output layer. Adam optimizer \cite{kingma2014adam} is applied to train the networks. 

\section{Experiments}
In this section, we experimentally evaluate our methods with the aim of answering the following questions: 
\begin{itemize}
\item[Q.1] Equipped with the learned IS strategies in LA, is the agent smart enough to hit the target entity?
\item[Q.2] Equipped with the learned KA strategies in LA, can the agent acquire knowledge rapidly to improve its knowledge base?
\item[Q.3] Is the acquired knowledge helpful to the agent in terms of increasing the efficiency in identifying the target entity?
\item[Q.4] Starting with no prior, can the logical connections between questions learned by the agent over episodes?  
\item[Q.5] Can the agent engage players in contributing their knowledge unknowingly by asking appropriate questions?
\end{itemize}

\subsection{Experimental Settings}
We evaluate our methods against the player simulators constructed from the real world data. Some important parameters are listed in Table \ref{tab:params LA-RL}.

\begin{table}[tbp]
\centering
\caption{Parameters for experiments.}
\label{tab:params LA-RL}
{\scalebox{.84}{
\begin{tabular}{c|cc}
\hline
Meth. & Parameter & Value \\
\hline
\multirow{8}{*}{{\rotatebox{90}{\small{LA-DQN \& LA-DRQN}}}}&learning rate & $2.5\times 10^{-4}$ \\
&replay prioritization  & $0.5$ \\ 
&initial explore  & $1$ \\
&final explore & $0.1$ \\
&final explore step  & $1\times 10^6$ \\
&discount factor ($\gamma$) & $0.99$ \\
&behave network update frequency in episodes& $1$ \\
&target network update frequency in episodes ($C$) & $1\times 10^4$ \\
&LA-DQN hidden layers' sizes on \textit{PersonSub}& [2048, 1024]\\
&LA-DRQN network sizes ($N_1$, $N_2$, $N_3$) on \textit{PersonSub}& [30, 2, 32] \\
&LA-DRQN network sizes ($N_1$, $N_2$, $N_3$) on \textit{Person}& [62, 2, 64] \\
\hline
\multirow{4}{*}{{\rotatebox{90}{\small{LA-GMF~}}}}&learning rate & $1\times 10^{-3}$ \\
&buffer size ($N_k$)& $9000$\\
&number of negative samples &  $4$\\
&dimension of latent factor ($K$) & $48$\\
\hline
\end{tabular}}}
\end{table}

\begin{figure}
\includegraphics[height=1.3in, width=2.1in]{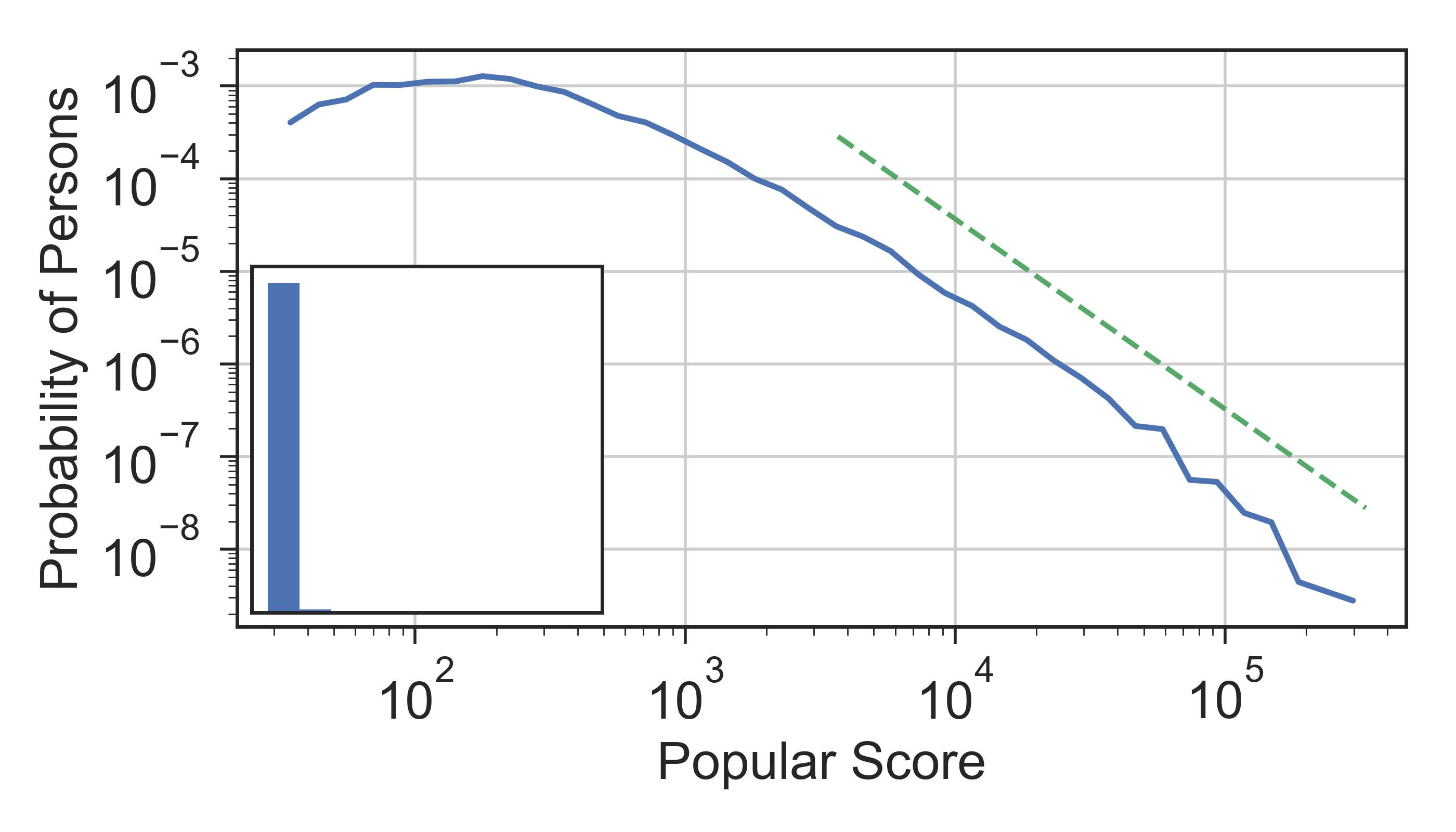}
\caption{Long tail distribution on log-log axes of Person.}
\label{fig:long tail entities}
\end{figure}

\textbf{Dataset.} We made an experimental Chinese \textit{20 Questions} system, where the target entities are persons (celebrities and characters in fictions), and collected the data for constructing the entity-question matrices. As described in Section 2, $\mathcal{R}=$ \{ \textit{yes, no, unknown} \} is the response set. Every entry follows a multinoulli distribution with corresponding collected response records. The entries with less than 30 records were filtered out, since they are statistically insignificant. For all missing entries, we initialized them with 3 records, one for each possible response, corresponding to a uniform distribution over \{\textit{yes, no, unknown}\}. After data processing, the whole dataset \textit{Person} contains more than 10 thousands entities with 1.8 thousands questions. Each entity has a popular score, indicating the frequency in which the entity appears in the game logs. The popular score follows a long tail distribution as shown in Figure \ref{fig:long tail entities}. For simplicity, we constructed a subset \textit{PersonSub} while comparing performances of different methods, which contains the most popular 500 entities and their most asked 500 questions. Both datasets summarized in Table \ref{tab:data summary} are in Chinese, where the missing ratio is the proportion of missing entries. The datasets are highly sparse with diverse questions, which makes it more challenging to learn better questioning strategies. It is worth emphasizing that the final datasets include only the aggregated entries instead of the session-level game logs, which means the agent has to figure out the questioning strategies interactively from zero.

\begin{table}[tbp]
\centering
\caption{Statistics of datasets.}
\label{tab:data summary}
{\scalebox{1}{
\begin{tabular}{cccc}
\hline
 Dataset & \# Entities & \# Questions & Missing Ratio \\
\hline
Person & 10440 & 1819 & 0.9597 \\
PersonSub & 500 & 500 & 0.5836 \\
\hline
\end{tabular}}}
\end{table}

\textbf{Simulator.} 
In order to evaluate our methods, a faithful simulator is necessary. The simulating approach not only obviates the expensive cost of online testing (It hurts players' game experience.) but also allows us to compare the acquired knowledge base with the ground-truth knowledge base which is not available while testing online. 
To simulate players' behavior, we consider two aspects: target generation and response generation. At the beginning of each episode, the player simulator samples an entity as the target according to the popular score distribution, which simulates how likely the agent would encounter that entity in a real episode. The response to a certain entity-question pair is obtained by sampling from the multinoulli distribution of the corresponding entry, which aims to model the uncertainty in a real player's feedback.

\textbf{Metrics.}
We use two metrics, one for IS evaluation, and the other for KA evaluation. (1) \textbf{Winning Rate.} To measure the agent's ability of hitting the target entity, we obtain the winning rate by averaging over a sufficient number of testing episodes where the agent uses the learned questioning strategy to play \textit{20 Questions} against the simulator. 
(2) \textbf{Knowledge Base Size vs. Distance.}
To evaluate the KA strategies, we consider not only the number of knowledge base entries that have been filled, i.e. knowledge base size, but also the effectiveness of the filled entries, given that fast knowledge base size increase may imply the agent wastes questioning opportunities on irrelevant questions and "unknown" answers. Also, an effective KA strategy is supposed to result in a knowledge base which reflects the knowledge across all the human players. In other words, the distance between the collected knowledge base and the human knowledge base should be small. Since each entry of the entity-question matrix is represented as a random variable with a multinoulli distribution in our methods, we define the asymmetric distance from knowledge base $D^1$ to $D^2$ as average exponential Kullback-Leibler divergence over all entries:
\begin{equation}
Distance(D^1, D^2) = \frac{\sum_{i,j} \exp\left({ \rm{KL}} (d^1_{ij}||d^2_{ij})\right)}{\hat{N}},
\end{equation}
where $\hat{N}$ is the number of total entries, $D^1$ and $D^2$ are the knowledge bases for the agent and the player simulator respectively. For computation convenience, the missing entries are smoothed as uniform distributions while calculating the distances. Clearly, a good KA strategy is expected to achieve fast knowledge base distance decrease while the knowledge base size increase slowly, which means the agent does not waste questioning opportunities on worthless questions.

\textbf{Baselines.} 
Since we study a novel problem in this paper, baseline methods are not readily available. Hence, in order to demonstrate the effects of our methods, we constructed baselines for each module separately by hand. Details about the baselines construction will be described later.

\subsection{Performance of IS Module (Q.1)}
In this subsection, we compare different methods to generate questioning strategies for the IS module. Our experiments on \textit{PersonSub} demonstrate the effectiveness of the proposed RL-style methods, and further experiments on \textit{Person} show the scalability of the proposed LA-DRQN. 

\begin{figure*}
\centering
\subfigure[]{\includegraphics[height=1.2in, width=1.7in]{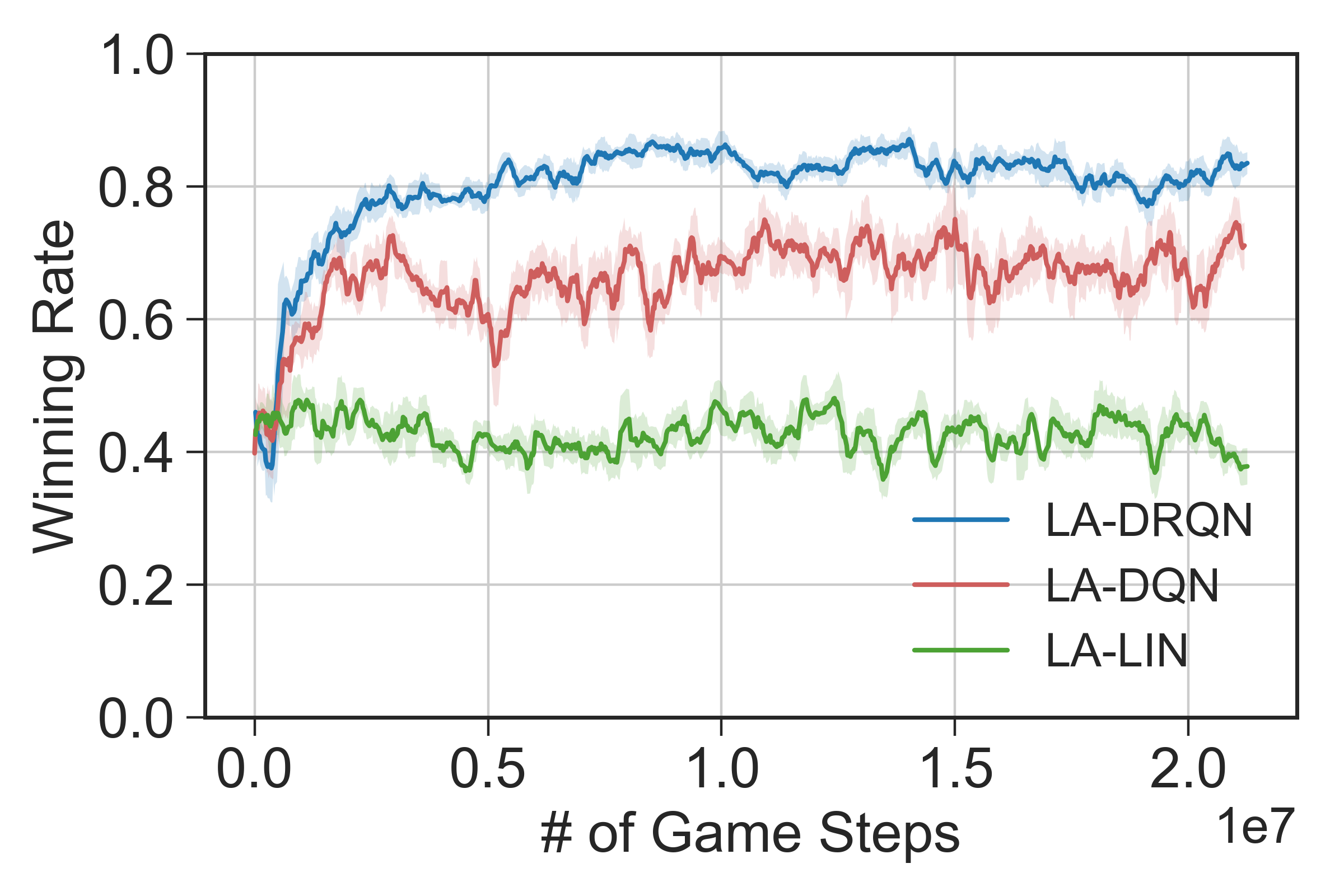}}
\subfigure[]{\includegraphics[height=1.2in, width=1.7in]{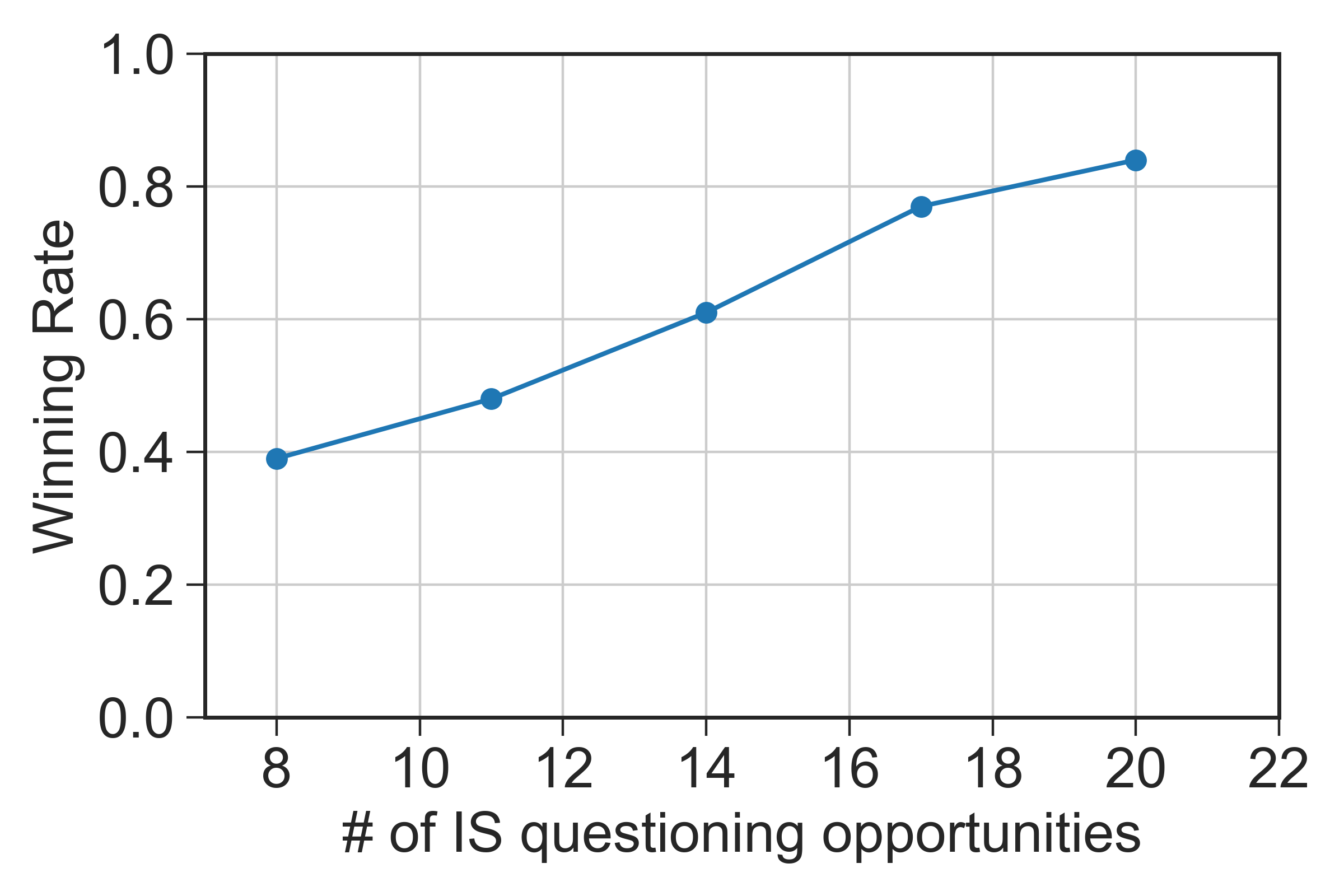}}
\subfigure[]{\includegraphics[height=1.2in, width=1.7in]{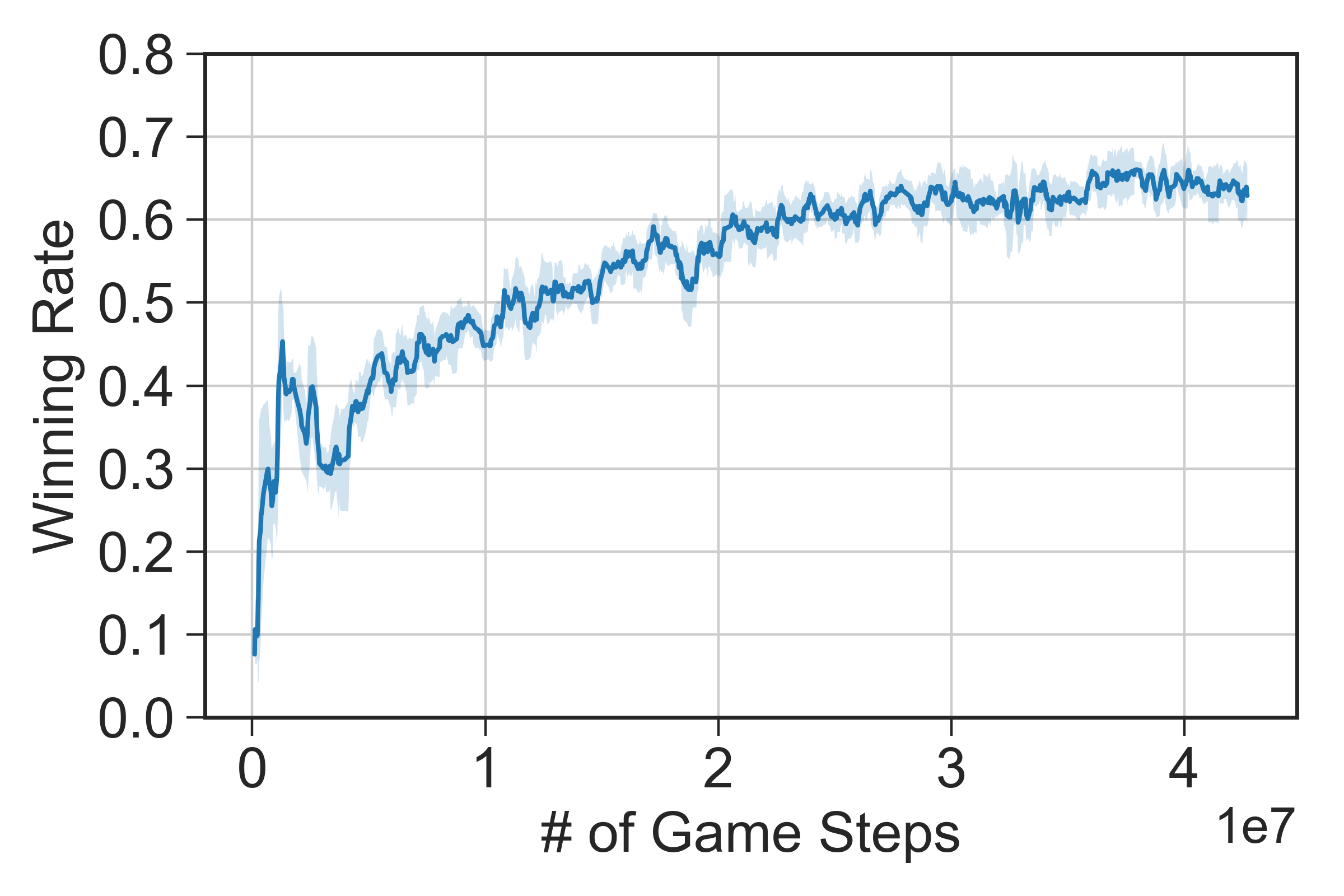}}
\subfigure[]{\includegraphics[height=1.2in, width=1.7in]{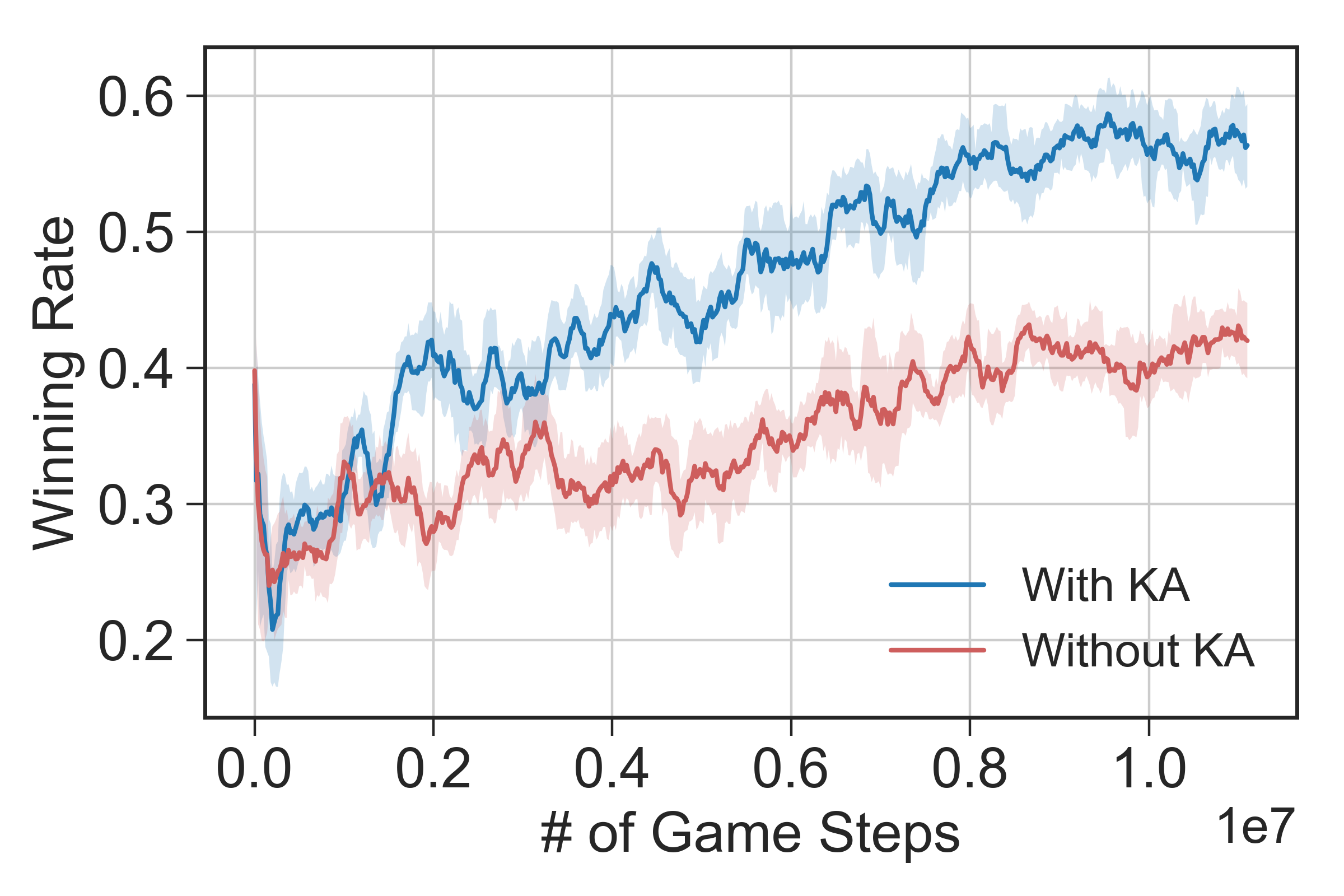}}
\caption{(a) Convergence processes on PersonSub; (b) LA-DRQN performance with varied IS questioning opportunities on PersonSub; (c) LA-DRQN convergence process on Person; (d) LA-DRQN performance of IS with KA on PersonSub.}
\label{fig:many}
\end{figure*}

\textbf{Baselines.} Methods based on entropy often play crucial roles in querying knowledge bases i.e. performing information-seeking tasks. Therefore, it is not surprising to find that the most popular approaches to build \textit{20 Questions} IS agent seem to be the fusion of entropy-based methods and heuristics (e.g.  asking the question with maximum entropy) as well. Although supervised methods like decision trees also work well as long as the game logs are available, these methods actually suffer from the cold-start problem, where the agent has nothing but a automatically constructed knowledge base. Given the fact that our primary goal is to complete a knowledge base which may be automatically constructed with no initial game logs available, we choose entropy-based methods as our baselines. We design several entropy-based methods, and take the best one as our baseline, named as Entropy. 
Specially, we maintain a candidate entity set ${\mathcal{E}}'$ initialized by $\mathcal{E}$. Every entity $e_m$ in ${\mathcal{E}}'$ is equipped with a tolerance value $t_m$ initialized by 0. Let $E_{mn}=\ep[d_{mn}]$ denote the expectation by setting $yes=1$, $no=-1$ and $unknown=0$. We define a variable $E_n$ for question $q_n$, which follows a distribution by considering all $\{E_{mn}|e_m \in {\mathcal{E}}'\}$. At each step, with the idea of maximum entropy, we choose the question $q_c={\arg\max}_{q_n} h(E_n)$ and receive the response $x_c$, where $h(\cdot)$ is the entropy function. Then for each $e_m\in {\mathcal{E}'}$, $t_m$ adds $|E_{mc}-x_c|$, which indicates the bias. When $t_m$ reaches a threshold (15 in our experiments), the entity $e_m$ will be removed from ${\mathcal{E}}'$. Finally, the entity with smallest tolerance value in ${\mathcal{E}}'$ is chosen to be the guess. Besides, we also design a linear Q-learning agent as a simple RL baseline, named as LA-LIN.

\textbf{Winning Rates.} Table \ref{tab:Performance of 20Q} shows the results for $T_1 = 20$. On PersonSub, the simple RL agent LA-LIN performs worse than the Entropy agent, suggesting that the Entropy agent is a relatively competitive baseline. Both the LA-DQN and the LA-DRQN agent learn better questioning strategies than the Entropy agent with higher winning rates achieved while playing \textit{20 Questions} against the player simulator. However, the winning rate of the LA-DRQN agent is significantly higher than that of the LA-DQN agent. 

\begin{table}[tbp]
\centering
\caption{Winning rate of the IS module. The reported results are obtained by averaging over 10000 testing episodes.}
\label{tab:Performance of 20Q}
{\scalebox{1}{
\begin{tabular}{ccccc}
\hline
 Dataset & Entropy & LA-LIN & LA-DQN & LA-DRQN \\
\hline
PersonSub & 0.5161 & 0.4020 & 0.7229 & 0.8535\\
Person & 0.2873 & $-$ & $-$ &0.6237\\
\hline
\end{tabular}}}
\end{table}

\textbf{Convergence Processes.} As shown in Figure \ref{fig:many}(a), with $T_1 = 20$, the LA-DRQN agent learns better questioning strategies faster than the LA-DQN agent due to the architecture of value network. The redundancy in states of the LA-DQN agent results in slow convergence and it takes much longer to train the large value network. In contrast, the embeddings and recurrent units in the LA-DRQN agent significantly shrink the sizes of the value network via extracting a dense representation of states from the complete asking and responding history. As for the LA-LIN, the model capacity is too small to capture a good strategy although it converges fast. 

\textbf{Number of Questioning Opportunities.} The number of questioning opportunities has considerable influence on the performance of IS strategies. Hence, we take a further look at the performance of the LA-DRQN agent trained and tested under varied number of questioning opportunities ($T_1)$. As shown in Figure \ref{fig:many}(b), where each point corresponds to the performance of the LA-DRQN agent trained up to $2.5\times 10^7$ steps, more questioning opportunities lead to higher winning rates. Given this fact, it is important to allocate enough questioning opportunities for the IS module if the agent wants to succeed in seeking the target i.e. winning the game. We observe that 17 questioning opportunities are adequate for the LA-DRQN agent to perform well on \textit{PersonSub}. Thus, the remaining 3 questioning opportunities can be saved for the KA module as will be discussed in detail later.

\textbf{Scale up to a Larger Knowledge Base.} Finally, we are interested in the scalability of our methods considering that our primary goal is to complete various knowledge bases. We perform experiments with $T_1 = 20$ on \textit{Person}, a larger knowledge base covering almost all data from the game system. It is worthwhile to mention that, \textit{20 Questions} becomes much more difficult for the agent due to the large sizes and high missing ratio of the entity-question matrix. Since the LA-LIN and LA-DQN give poor results due to huge state spaces (The number of the dimensions of the states is 40k on \textit{Person}.), and the computation complexity grows linearly with the number of questions, they are difficult to scale up to larger spaces of questions and responses. Hence, we only evaluate the Entropy baseline and the LA-DRQN method. The winning rates are shown in Table \ref{tab:Performance of 20Q} while the convergence process of LA-DRQN is presented in Figure \ref{fig:many}(c). The LA-DRQN agent outperforms the Entropy agent significantly, which demonstrates its effectiveness in seeking the target and its ability to scale up to larger knowledge bases.

\subsection{Performance of KA Module (Q.2)}
A promising IS module is the bedrock on which the KA module is based. Therefore, we choose the LA-DRQN, which outperforms all other methods, to instantialize the IS module while running KA experiments. We build a simulation scenario where the players own more knowledge than the agent and the agent tries to acquire knowledge via playing \textit{20 Questions} with the players. We ran experiments on \textit{PersonSub}. Note that in our KA experiments, the agent has $T_1 = 17$ questioning opportunities for IS and $T_2 = 3$ questioning opportunities for KA each episode, as mentioned in the study on the number of questioning opportunities.

\textbf{A Simulation Scenario for KA.} 
To build a simulation scenario for KA, we first randomly throw away 20\% existing entries of the original entity-question matrix and use the remaining dataset to construct the knowledge base for our agent. Throughout the whole interactive process, the knowledge base for the player simulator remains unchanged while the agent's knowledge base improves gradually and converges to the one the player simulator possesses theoretically in the end. 

\textbf{Baseline.} Since KA methods depend on the exact approach in which the knowledge is represented, it is difficult to find readily available baselines. However, we believe that all KA methods should take into account the "uncertainty" and "value" as Algorithm \ref{alg:ka} does, even though the "uncertainty" and "value" can be defined in different ways. Given this fact, reasonable baselines for KA should include the uncertainty-only method and the value-only method. For this reason, we simply modify Algorithm \ref{alg:ka} to obtain the baselines.

\begin{figure}
\centering
\subfigure[]{\includegraphics[height=1.14in, width=1.45in]{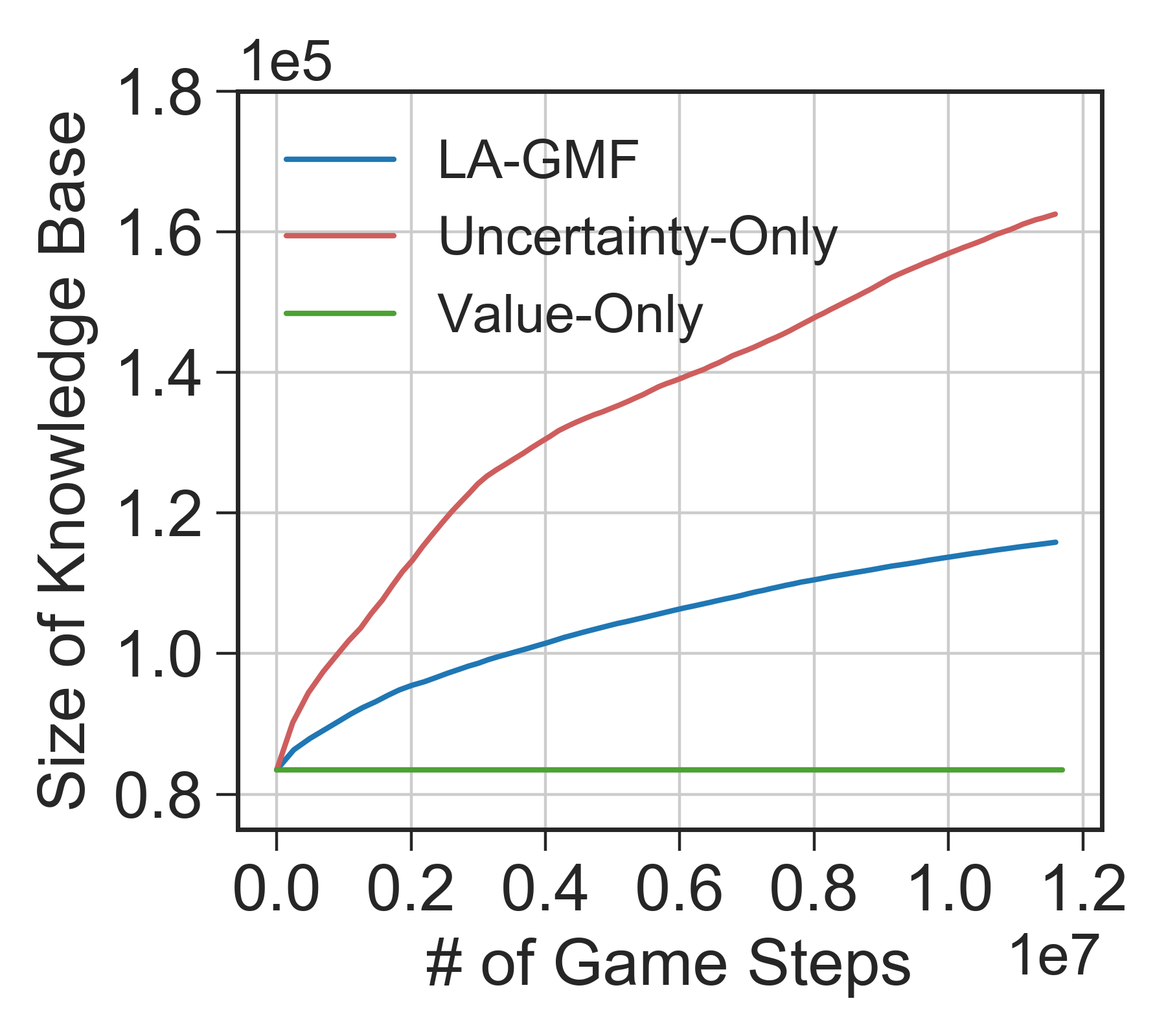}}
\subfigure[]{\includegraphics[height=1.12in, width=1.45in]{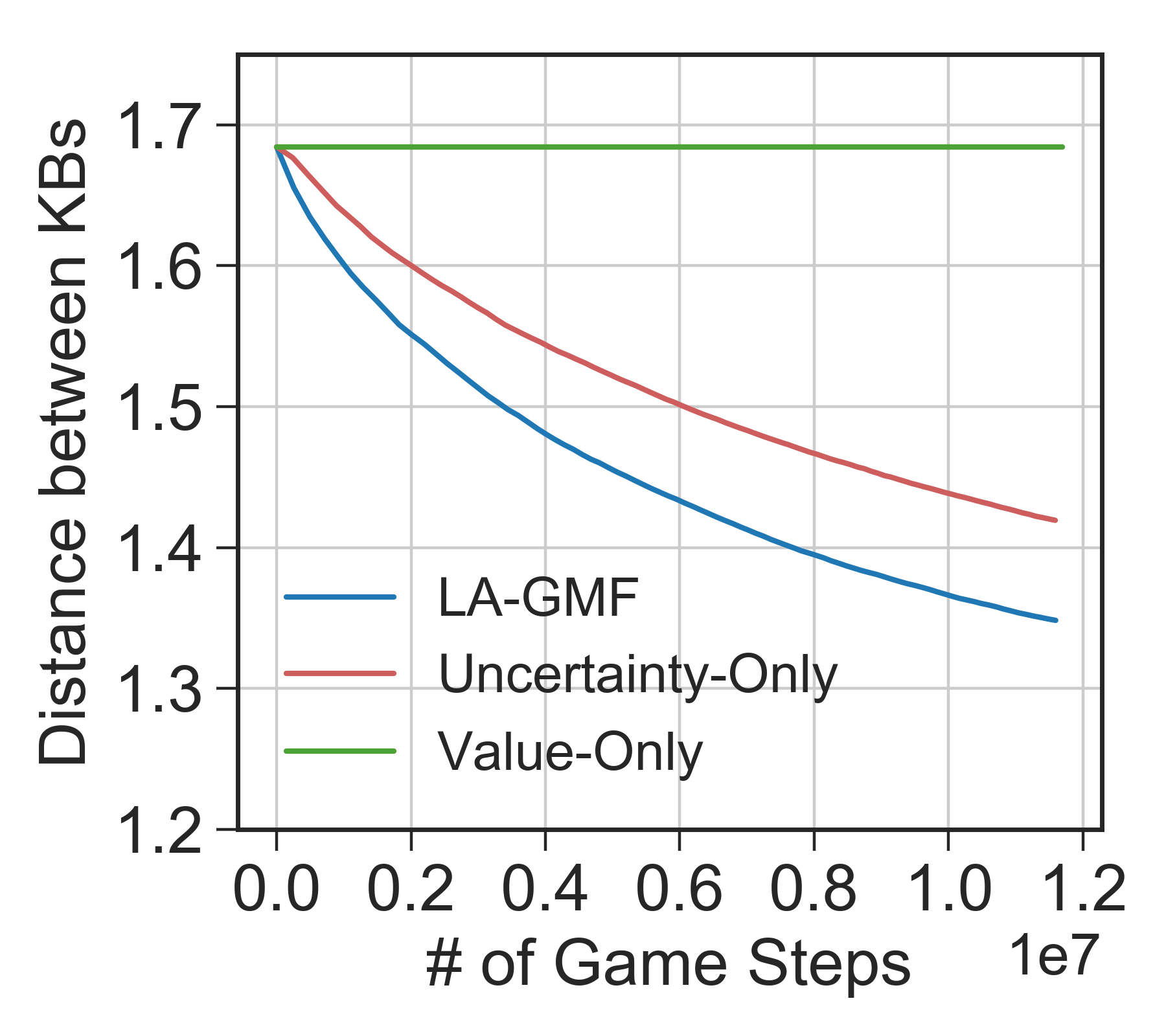}}
\caption{Knowledge base (a) size vs. (b) distance of agent on \textit{PersonSub}. The LA-GMF was trained with $N_c = 32$. }
\label{fig:ka}
\end{figure}

\textbf{Knowledge Base Size vs. Distance.} Figure \ref{fig:ka} shows the change of knowledge base sizes and knowledge base distances. The value-only method overexploits the value information estimated from the known entries, so the agent tends to ask more about the known entries instead of the unknown entries, leading to the almost unchanged knowledge base size and knowledge base distance. While the uncertainty-only method does collect new knowledge, it adopts a relatively inefficient KA strategy, wasting a lot of questioning opportunities on the nonsense knowledge (unknown by the players). That is why the knowledge base distance decreases quite slowly when the knowledge size soars up. Compared to the uncertainty-only method, our proposed LA-GMF enjoys a lower speed of knowledge base size increment and a higher speed of knowledge base distance reduction, indicating the agent learns to use the questioning opportunities in a more efficient way. To sum up, the results demonstrate that our proposed LA-GMF learns better strategies to acquire valuable knowledge faster and the knowledge base quickly becomes close to the ground-truth one (that of the player simulator).

\begin{figure}
\centering
\subfigure[]{\includegraphics[height=1.14in, width=1.45in]{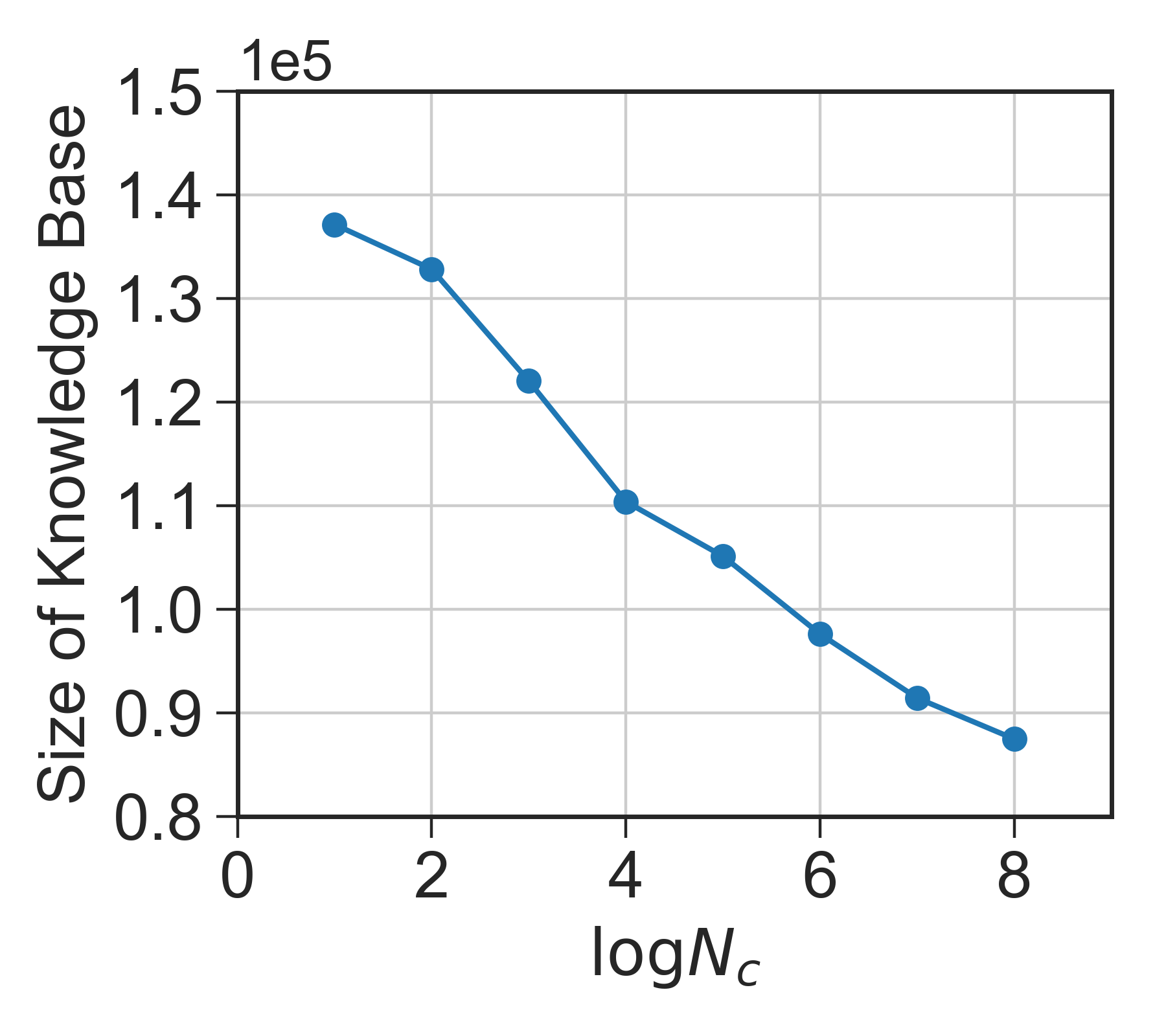}}
\subfigure[]{\includegraphics[height=1.12in, width=1.45in]{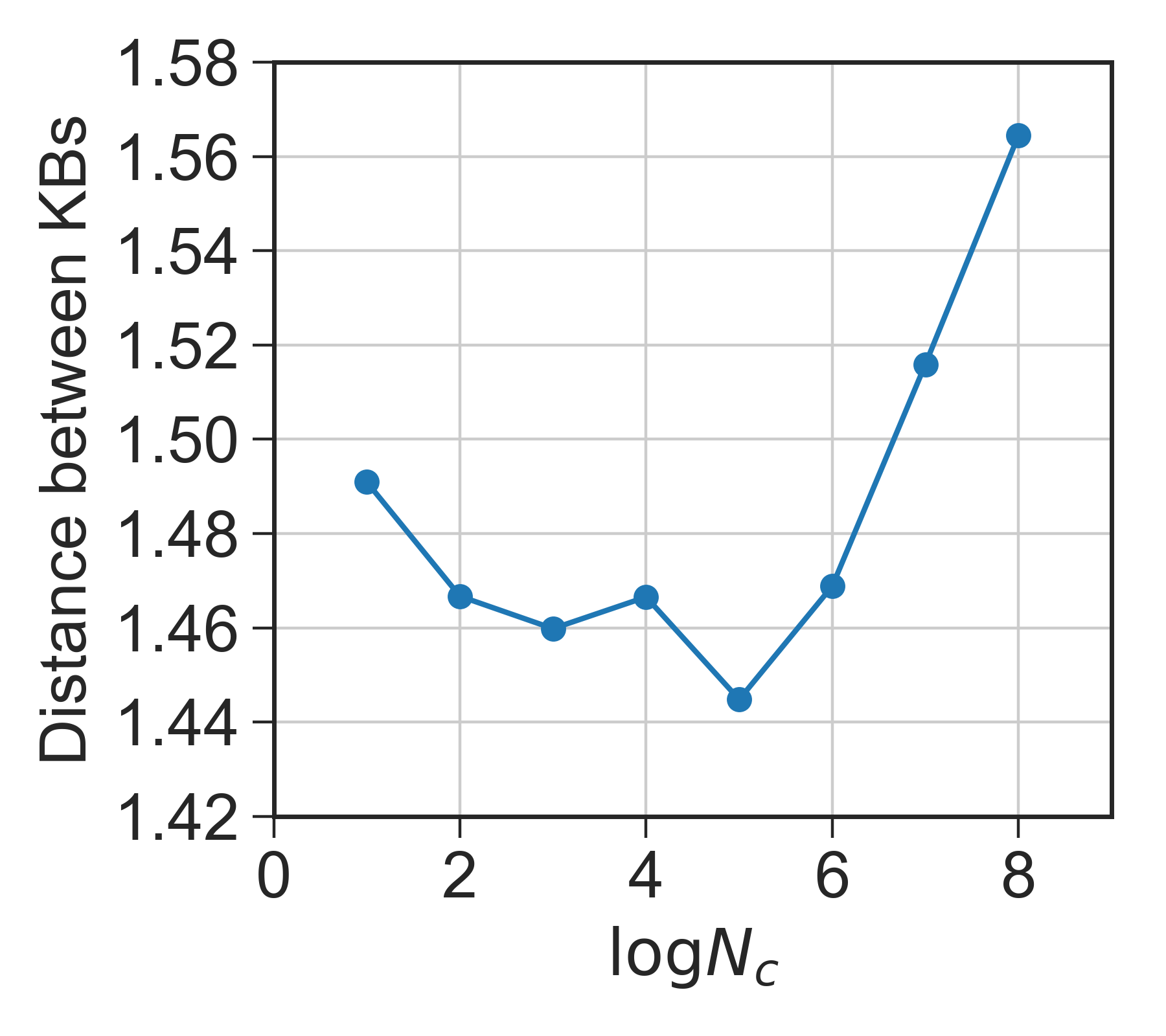}}
\caption{Tradeoff "uncertainty" and "value" on \textit{PersonSub}.}
\label{fig:ka_tradeoff}
\end{figure}

\textbf{Trade-off Uncertainty and Value.} The LA-GMF method consider both the "uncertainty" and "value". Actually, the parameter $N_c$ in Algorithm \ref{alg:ka} determines the trade-off between the two aspects, with larger $N_c$ putting more weight on "value" and smaller $N_c$ putting more weight on "uncertainty". Results of experiments on \textit{PersonSub} with different $N_c$ are shown in Figure \ref{fig:ka_tradeoff}. Each point in the figure was obtained by choosing the latest update in the previous 5.5 million game steps. In our experiments, $N_c = 32$ achieves the most promising knowledge base distance reduction with modest knowledge base size increment.

\subsection{Further Analysis}

\subsubsection{IS Boosted by the Acquired Knowledge (Q.3)} With the right guess in IS module, knowledge can be acquired via KA module. In return, the valuable acquired knowledge boosts the winning rates of the IS module. We take a close look at the effect of the acquired knowledge on IS. As shown in Figure \ref{fig:many}(d), our LA-DRQN agent that updates the IS strategy with the acquired knowledge performs better than the one that updates the IS strategy without the acquired knowledge, improving the winning rate faster i.e. learning better IS strategies. 

\subsubsection{Learned Question Representations (Q.4)}
\begin{figure*}
\includegraphics[height=1.3in, width=5.8in]{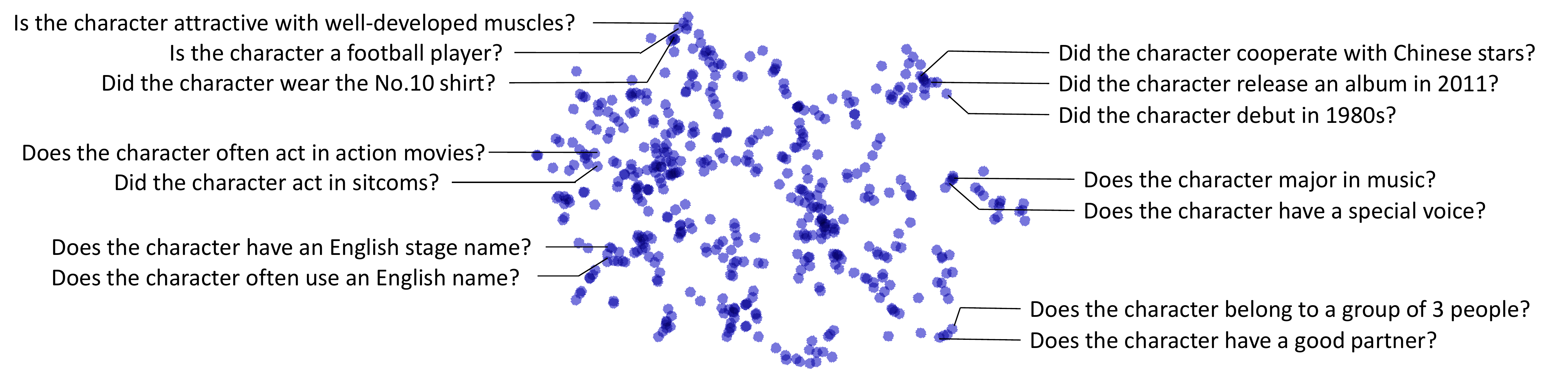}
\caption{Visualization of question embeddings with LA-DRQN on PersonSub.}
\label{fig:embedding}
\end{figure*}

To analyze whether our LA-DRQN agent can learn reasonable question representations, we visualize the learned question embeddings $W_1(q_n)~(n=1,2,...,N)$ on \textit{PersonSub} dataset using t-SNE. We annotate several typical questions in Figure \ref{fig:embedding}. It can be observed that if questions have logical connections between each other, they may have similar embeddings i.e close to each other in the embedding space. For example, the logical connected questions "Is the character a football player?" and "Did the character wear the No.10 shirt?" are close to each other in the learned embedding space. This suggests that we could obtain meaningful question embeddings starting from random initialization after the agent plays a sufficient number of episodes of the interactive \textit{20 Questions}, although initializing the question embeddings with pre-trained word embeddings would bring faster convergence in practice.

\subsubsection{Questioning Strategy for KA (Q.5)}
To intuitively analyze the knowledge acquisition ability of our agent, we show several contrastive examples of KA questions generated by LA-GMF and uncertainty-only method in Table \ref{tab:ka example}. As observed, the questions generated by LA-GMF are more reasonable and relevant to the target entity, while the ones generated by uncertainty-only method are incongruous and may reduce players' gaming experience. Moreover, the LA-GMF agent tends to ask valuable questions that players know the answers and acquire valid facts instead of getting "\textit{unknown}" and wasting the opportunities. For example, with the target entity Jimmy Kudo, the LA-GMF agent chooses to ask questions like "Is the character brave?" to obtain the fact "Jimmy Kudo is brave", while the uncertainty-only agent chooses some strange questions like "Is the character a monster?" and wastes the valuable questioning opportunities.

\begin{table*}[tbp]
\centering
\caption{Contrasting examples of KA questions generated by LA-GMF and Uncertainty-Only.}
\label{tab:ka example}
{\scalebox{1}{
\begin{tabular}{c|cc|cc}
\hline
Target Entity & \multicolumn{2}{c|}{KA questions with LA-GMF} & \multicolumn{2}{c}{KA questions with Uncertainty-Only} \\
\hline
\multirow{3}{*}{ Jay Chou\footnotemark}&Does the character mainly sing folk songs? & No & Did the character die unnaturally? & Unknown\\
&Is the character handsome? & Yes& Does the character like human beings?& Unknown \\ 
&Is the character born in Japan? & No & Is the character from a novel?& No \\
\hline
\multirow{3}{*}{ Jimmy Kudo\footnotemark}&Is the character brave? & Yes & Does the character only act in movies (no TV shows)? & Unknown\\
&Is the character from Japanese animation? & Yes& Has the character ever been traitorous?& Unknown \\ 
&Was the character dead? & No & Is the character a monster?& Unknown \\
\hline
\end{tabular}}}
\end{table*}

\section{Conclusion and future work}
In this paper, we study knowledge acquisition via \textit{20 Questions}. We propose a novel framework, \textit{learning-to-ask} (LA), within which automated agents learn to ask questions to solve an information seeking task while acquiring new knowledge within limited questioning opportunities. We also present well-designed methods to implement the LA framework. Reinforcement learning based methods, LA-DQN and LA-DRQN, are proposed for information seeking. Borrowing ideas from recommender systems, LA-GMF is proposed for knowledge acquisition. Experiments on real data show that our agent can play enjoyable \textit{20 Questions} with high winning rates, while acquiring knowledge rapidly for knowledge base completion. 

In fact, LA framework can be implemented in many other approaches. Our work just marks a first step towards deep learning and reinforcement learning based knowledge acquisition in a game-style Human-AI collaboration. We believe it is worthwhile to undertake in-depth studies on such smart knowledge acquisition mechanisms in an era where various Human-AI collaborations are becoming more and more common.

For future work, we are interested in integrating the IS and KA seamlessly using hierarchical reinforcement learning. And a bandit-style KA module is worth studying since the KA essentially trades off the "uncertainty" and "value" similar as in the exploitation-exploration problem. Furthermore, modeling the complex dependencies between questions explicitly should be helpful to the IS. 

\footnotetext[3]{Jay Chou is a famous Chinese singer.(https://en.wikipedia.org/wiki/Jay\_Chou)}
\footnotetext[4]{Jimmy Kudo is a role in a Japanese animation.(https://en.wikipedia.org/wiki/Jimmy\_Kudo)}

\begin{acks}
This work is partly supported by the National Natural Science Foundation of China (Grant No. 61673237 and No. U1611461). Thank reviewers for their constructive comments.
\end{acks}

\bibliographystyle{ACM-Reference-Format}
\bibliography{bibliography-LA}

\end{document}